# Modeling arousal potential of epistemic emotions using Bayesian information gain: Inquiry cycle driven by free energy fluctuations

Hideyoshi Yanagisawa[*a], Shimon Honda[b]

[a] The University of Tokyo, 7-3-1 Bunkyo, Hongo, Tokyo, 113-8656, JAPAN. hide@mech.t.u-tokyo.ac.jp

[b] The University of Tokyo, 7-3-1 Bunkyo, Hongo, Tokyo, 113-8656, JAPAN. hondar8@g.ecc.u-tokyo.ac.jp

*Corresponding author

## Abstract

Epistemic emotions, such as curiosity and interest, drive the inquiry process. This study proposes a novel formulation of epistemic emotions such as curiosity and interest using two types of information gain generated by the principle of free energy minimization: Kullback–Leibler divergence (KLD) from Bayesian posterior to prior, which represents free energy reduction in recognition, and Bayesian surprise (BS), which represents the expected information gain by Bayesian prior update. By applying a Gaussian generative model with an additional uniform likelihood, we found that KLD and BS form an upward-convex function of surprise (minimized free energy and prediction error), similar to Berlyne's arousal potential functions, or the Wundt curve. We consider that the alternate maximization of BS and KLD generates an ideal inquiry cycle to approach the optimal arousal level with fluctuations in surprise, and that curiosity and interest drive to facilitate the cyclic process. We exhaustively analyzed the effects of prediction uncertainty (prior variance) and observation uncertainty (likelihood variance) on the peaks of the information gain function as optimal surprises. The results show that greater prediction uncertainty, meaning an open-minded attitude, and less observational uncertainty, meaning precise observation with attention, are expected to provide greater information gains through a greater range of exploration. The proposed mathematical framework unifies the free energy principle of the brain and the arousal potential theory to explain the Wundt curve as an information gain function and suggests an ideal inquiry process driven by epistemic emotions.

## 1. Introduction

Inquiry is an essential cognitive process in human activities such as scientific research, creation, and education. American philosopher Charles Sanders Peirce defines inquiry as a cycle of three inferences: abduction, deduction, and induction (Peirce, 1974). In the observation of surprising phenomena, abduction infers a possible cause of the observation, deduction predicts unknown effects based on the inferred cause, and induction tests the prediction and updates the causal knowledge. A voluntary inquiry process is facilitated by epistemic emotions such as surprise, curiosity, interest, and confusion (Kashdan & Silvia, 2009; Vogl, Pekrun, Murayama, & Loderer, 2020). Psychologist Berlyne defined two types of epistemic curiosity: diversive and specific (Berlyne, 1966; Silvia, 2012). Diversive curiosity seeks novelty, and thus, in this type of curiosity, surprise triggers abductive reasoning. On the other hand, specific curiosity drives induction, which seeks evidence of deductive reasoning to resolve confusion.

Emotions are generally mapped to a dimensional space (Lang, 1995; Russell, 1980). The most commonly used dimensions are arousal and valence, termed the *core affect* (Russell, 2003). Arousal is the intensity of emotions, whereas valence is the dimension of the positive and negative poles. A recent functional magnetic resonance imaging (fMRI) study showed that arousal and valence are correlated with neural activity in the orbitofrontal cortex and amygdala, respectively (Wilson-Mendenhall, Barrett, & Barsalou, 2013). The emotional dimensions are not independent, and arousal affects valence. Berlyne's arousal potential theory suggests that an appropriate level of arousal potential induces a positive hedonic response, whereas extreme arousal induces a negative response (Berlyne, 1960). Thus, valence forms an inverse-U-shaped function of the arousal potential, termed the *Wundt curve* (Fig. 1). Berlyne suggests that epistemic curiosity approaches the optimal arousal potential, where the hedonic response (or positive valence) is maximized (Berlyne, 1960, 1966; Silvia, 2012).

Berlyne also illustrated a number of arousal potential factors such as novelty, complexity, and uncertainty (Berlyne, 1960). Yanagisawa mathematically explains that the *free energy*, which is information on the brain's prediction error or *surprise* (Friston, Kilner, & Harrison, 2006), represents the arousal potential because free energy is decomposed into information quantity terms representing perceived novelty, complexity, and uncertainty (Yanagisawa, 2021). This free-energy arousal model suggests that an appropriate level of free energy or surprise induces a positive emotional valence based on Berlyne's Wundt curve, which is supported by experimental evidence (Honda, Yanagisawa, & Kato, 2022; Sasaki, Kato, & Yanagisawa, 2023).

By contrast, the free energy principle (FEP) (Friston et al., 2006), known as the unified brain theory (Friston, 2010), suggests that the brain must minimize its free energy during perception and action. Previous studies have proposed that decreasing and increasing free energy (or expected free energy) correspond to positive and negative valence, respectively (Clark, Watson, & Friston, 2018; Hesp et al., 2021; Joffily & Coricelli, 2013; Seth & Friston, 2016; Wager et al., 2015; Yanagisawa, Wu, Ueda, & Kato, 2023), and that high and low free energies indicate uncertain and certain states, respectively. Reducing free energy resolves uncertainty and produces positive emotions.

The FEP argument that minimizing free energy corresponds to a positive valence seems to contradict the argument of arousal potential theory that an appropriate level of arousal potential (represented by free energy (Yanagisawa, 2021)) maximizes positive valence. To resolve this contradiction and integrate the FEP-based valence and arousal potential theories, we propose a novel valence framework based on the theory that a decrement in free energy and its expectation explain the valence of epistemic emotions. A decrease in free energy represents information gain and an epistemic value (Friston et al., 2017; Parr, Pezzulo, & Friston, 2022). The more information gain (epistemic value) one obtains or expects, the more positive the valence one experiences.

Based on this framework, we formulated emotion valence functions of the arousal potential using decrements in free energy (or information gains). By applying a Gaussian generative model with an additional uniform likelihood, we demonstrated that the epistemic valence function forms an inverse-U shape and analyzed the effects of prediction error and uncertainties on the peaks of the valence functions. We associated epistemic emotions such as curiosity and interest with the free-energy-based valence model. Furthermore, we proposed an inquiry cycle model based on free-energy-based epistemic emotions.

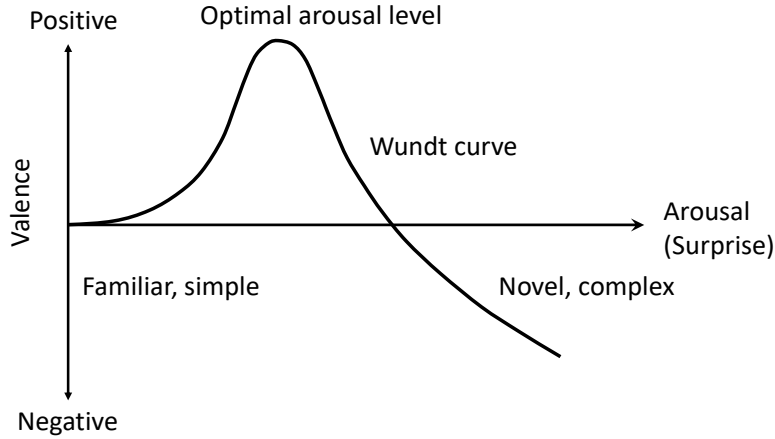

Fig. 1 Arousal potential function, or Wundt curve. Appropriate level of arousal maximizes positive emotion valence (optimal arousal level).

## 2. Method

### 2.1 Free energy formulations

FEP suggests that the brain must minimize its free energy through recognition, action, and learning (Friston et al., 2006). Assume an agent recognizes a hidden state $s$ as a cause of an observation $o$ given by an action based on a policy $\pi$. We assume that the agent has a generative model $p(s, o|\pi)$ as its knowledge about the probabilistic relationship between hidden states and observation and a recognition density $q(s|\pi)$ of hidden states for a given policy. The free energy of a policy $\pi$ is defined as a function of an observation representing the difference between a recognition density and a generative model averaged by the recognition density in terms of their energies (negative log probability).

$$F_\pi = \langle \ln q(s|\pi) - \ln p(s, o|\pi) \rangle_{q(s|\pi)} \tag{1}$$

The free energy represents the prediction error of recognition from the knowledge, i.e., the generative model. It refers to uncertainty and the prediction error of signals in a Bayesian brain theory (Knill & Pouget, 2004). The first and second terms on the right-hand side denote the negative-state entropy and internal energy, respectively. Thus, the definition corresponds to the Helmholtz free energy when the temperature is one.

With the definition of conditional probability, the generative model is factorized into true posterior and evidence: $p(s, o|\pi) = p(s|o, \pi)p(o|\pi)$. With this factorization, the free energy is expanded to the summation of a Kullback–Leibler (KL) divergence and Shannon surprise (hereafter referred to as *surprise*).

$$F_\pi = D_{KL}[q(s|\pi)||p(s|o, \pi)] - \ln p(o|\pi) \tag{2}$$

The first-term KL divergence forms the true posterior to the recognition density, which represents a statistical difference between the two distributions: $D_{KL}[q(s|\pi)||p(s|o, \pi)] = \langle \ln q(s|\pi) - \ln p(s|o, \pi) \rangle_{q(s|\pi)}$. When the recognition approximates the true posterior to minimize free energy, the KL divergence becomes zero, and the free energy is approximated to the second term, i.e., surprise. Thus, the lower bound of free energy is surprise. Surprise is a negative log of the model evidence, $p(o|\pi)$, and refers to the information content used to process given observations, representing cognitive load (Yanagisawa, 2021).

The generative model is decomposed to a state prior $p(s|\pi)$ for a given policy and a likelihood function $p(o|s)$.

$$p(s, o|\pi) = p(s|\pi)p(o|s) \tag{2}$$

With this decomposition, the free energy is expanded to another two terms.

$$F_\pi = D_{KL}[q(s|\pi)||p(s|\pi)] - \langle \ln p(o|s) \rangle_{q(s|\pi)} \tag{3}$$

The first term is a KL divergence of state prior from recognition. This term represents the complexity of the generative model. The second term is the difference between likelihood and recognition. This term indicates negative model accuracy. Thus, minimizing the free energy signifies minimizing the complexity and maximizing the accuracy of the model.

## 2.2 Information gain in recognition

Assume that an initial recognition density before an action based on a policy $\pi$ is approximated to the state prior. The initial free energy $F_{\pi 0}$ is a summation of KL divergence and surprise.

$$F_{\pi 0} = \langle \ln p(s|\pi) - p(s, o|\pi) \rangle_{p(s|\pi)} = D_{KL}[p(s|\pi)||p(s|o, \pi)] - \ln p(o|\pi) \tag{4}$$

An agent receives an observation $o$ by the action based on the policy $\pi$. The recognition density approximates the true posterior by minimizing the free energy. The KL divergence becomes zero, and the free energy decreases to the lower bound $F_{\pi R}$, corresponding to surprise.

$$q(s|\pi): p(s|\pi) \to p(s|o, \pi), \; F_{\pi R} = -\ln p(o|\pi) \tag{5}$$

The decrease in free energy in the recognition process is equivalent to the KL divergence from the true posterior to the initial recognition, $KLD_\pi$. Herein, $KLD_\pi$ denotes the information gain from recognizing the causal state of observations given by an action based on a policy $\pi$.

$$\Delta F_R = F_{\pi 0} - F_{\pi R} = KLD_\pi = D_{KL}[p(s|\pi)||p(s|o, \pi)] \tag{6}$$

A greater KLD indicates that the recognition of an observation under a policy provides greater information gain. Thus, KLD represents the epistemic value of recognizing an observation under a policy. This suggests that an agent prefers to recognize observations with a greater KLD and is motivated to act based on a policy that likely obtains such observations. Therefore, we infer that KLD increases positive valence by increasing information gain (epistemic value) in recognition.

## 2.3 Information gain expected from Bayesian updating prior belief: Bayesian surprise

The free energy minimized by a recognition, $F_{\pi R}$, approximates surprise. The minimized free energy equals a summation of complexity and inverse accuracy with a recognition approximated to the true posterior, $q(s|\pi) \approx p(s|o, \pi)$.

$$F_{\pi R} = -\ln p(o|\pi) = BS_\pi + U_\pi \tag{7}$$

$$BS_\pi = D_{KL}[p(s|o, \pi)||p(s|\pi)] \tag{8}$$

$$U_\pi = -\langle \ln p(o|s) \rangle_{p(S|O, \pi)} \tag{9}$$

The complexity and inverse accuracy terms represent the Bayesian surprise $BS_\pi$ and perceived uncertainty $U$, respectively, and their summation (surprise) denotes the arousal potential (Yanagisawa, 2021). The Bayesian surprise, $BS_\pi$, is a KL divergence from posterior to prior, i.e., the deviation of recognition from prior expectation. It represents the novelty of the recognized observation and is correlated with the surprise response to novel stimuli (Yanagisawa, Kawamata, & Ueda, 2019). The surprise response decreases with repeated exposure to the same novel stimuli. Such habituation is formulated as a decrease in BS in the Bayesian update of the prior (Ueda, Sekoguchi, & Yanagisawa, 2021).

By repeatedly observing the same observation $o$ by an action under the same policy, the prior is updated by Bayesian updating such that the prior comes close to the true posterior, i.e., $p(s|\pi): p(s|\pi) \rightarrow p(s|o, \pi)$. When the prior is updated to the posterior, $BS_\pi$ is zero, and the free energy decreases to the inverse accuracy term. We refer to this term as *uncertainty* because it refers to the perceived uncertainty (Yanagisawa, 2021). Thus, the lower bound of free energy after the prior updating is the uncertainty, $F_{\pi L} \leq U_\pi$, whereas the upper bound of the free energy decrease is the Bayesian surprise, $BS_\pi$.

$$\Delta F_L = F_{\pi R} - F_{\pi L} \leq BS_\pi \tag{10}$$

Herein, $BS_\pi$ is equivalent to the maximum information gain expected from the prior update based on observing a sufficient number of the same observations given under the same policy. A greater $BS_\pi$ denotes a greater information gain expected from the update with the action under the policy $\pi$. Thus, BS represents the expected epistemic value given by the model (prior) update or learning. This suggests that an agent prefers novel observations with a greater BS, which is expected to provide a chance to learn new information (update its own generative model), and that the agent is motivated to approach such novel observations. Therefore, we infer that BS increases emotional valence in anticipation of information gain from updating prior beliefs.

## 2.4 Linking free energy reduction, information gain, and arousal potential

Fig. 2 summarizes the two-step free energy reduction and information gain. The free energy given an observation $o$ decreases by $KLD$ as the first information gain when one succeeds in recognizing the state as a cause of the observation. The minimized free energy approximates surprise. The surprise is a summation of $BS$ and $U$. When one's prior is updated to approximate the true posterior, the free energy is decreased by $BS$, which is the expected second information gain.

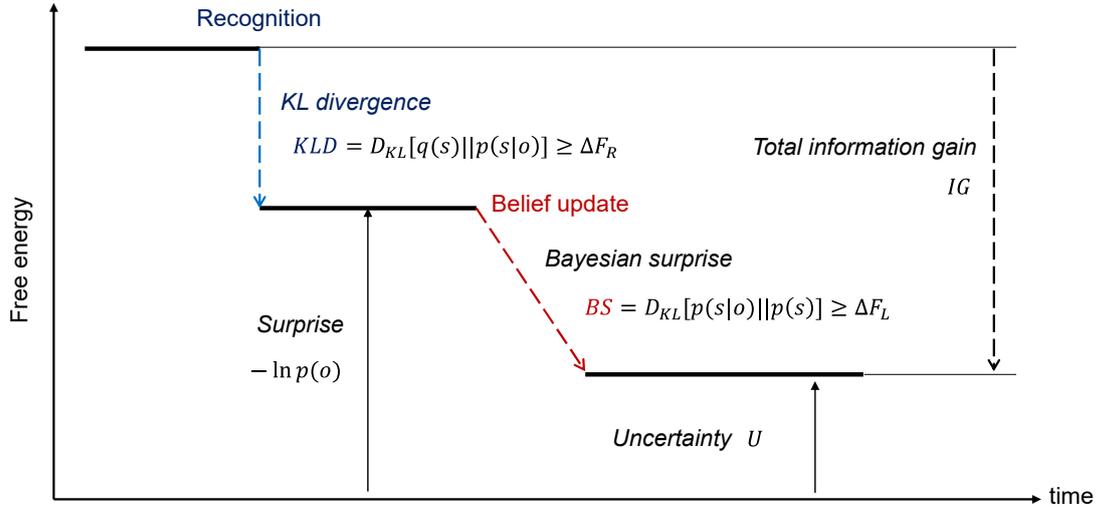

Fig. 2 Two-step free energy reduction and information gain. Decreases in free energy in recognition and belief update correspond to KL divergence (KLD) and Bayesian surprise, respectively.

The upper bound of the total free energy reduction (or information gain) from recognizing and updating state beliefs, given an observation, is a summation of the two KL divergences, i.e., information gain.

$$\Delta F_R + \Delta F_L \leq KLD_\pi + BS_\pi = D_{KL}[p(s|\pi)||p(s|o,\pi)] + D_{KL}[p(s|o,\pi)||p(s|\pi)] \equiv IG$$

(11)

We consider that the total information gain represents the epistemic values that explain the emotional valence of the arousal potential.

The two types of KL divergence denote the difference between the prior and posterior. When the posterior given an observation is the same as the prior, the KL divergences are zero, and the observation provides minimum free energy and minimum surprise (or maximum evidence). Hence, an observation that provides minimal free energy does not provide any KL divergence or information gain. To provide epistemic value with an emotional valence, given information gain, a certain level of surprise representing arousal potential (Yanagisawa, 2021) is required by observing unexpected outcomes that give certain KL divergences. However, if the likelihood of an observation is far from the prior distribution, where the likelihood does not provide any information, the posterior is not updated from the prior. In this case, the KL divergences are zero, and the observation does not provide any information. Therefore, we consider that an appropriate level of surprise maximizes the KL divergences (information gains) and that an appropriate level represents the optimal arousal potential that maximizes the positive valence for its epistemic value.

KL divergence is an asymmetric operation. Hence, although both KL divergences, $KLD_\pi$ and $BS_\pi$, denote differences between the prior and posterior, they are different from each other. This suggests that the two KL divergences as functions of surprise are different. KLD signifies information gain due to recognition, whereas BS signifies information gain expected from updating prior beliefs. Namely, KLD is the current, whereas BS is the future, information gain. This suggests that maximizing KLD and BS are different strategies for approaching the optimal arousal level that maximizes the total epistemic value with a positive valence.

## 2.5 Analytical methodology

We modeled the two information gains, KLD and BS, as functions of surprise using a Gaussian-like generative model with a flat likelihood of uniform noise and demonstrated that the two functions, KLD and BS, form an inverse-U shape and have different peaks. Using the function model, we analyzed the effect of Gaussian parameters, the difference between the prior mean and likelihood peak as *prediction error* (Yanagisawa, 2016), variance of prior as *prediction uncertainty*, and variance of likelihood as *observation uncertainty* on the peaks of the information gain functions. From the analysis, we elucidated the conditions for optimal prediction errors and uncertainties of prediction and observation to maximize the information gains in an ideal inquiry process.

## 3. Results

### 3.1 Gaussian model of information gains

The Gaussian Bayesian model has been used in past research studies to analyze the characteristics of free energy and Bayesian surprise (Buckley, Kim, McGregor, & Seth, 2017; Yanagisawa et al., 2023; Yanagisawa, 2021). The Laplace approximation suggests that a Gaussian distribution is applied around the mode of unknown distributions. The Gaussian form is useful for analyzing the effect of interpretable and independently manipulatable parameters on free energy and KL divergence. The distance between the prior mean and likelihood peak, $\delta$, represents *prediction error*; the variance of prior, $s_p$, represents *prior uncertainty*; and the variance of likelihood, $s_L$, represents *observation uncertainty*. The likelihood function of $n$ data randomly sampled from a source following a Gaussian distribution is

$$p(o^n|s) = \left(\frac{1}{\sqrt{2\pi s_L}}\right)^n \exp\left[\frac{-(s-\bar{o})^2 + nV}{2s_L}\right], \tag{12}$$

where $\bar{o}$ and $V$ denote the mean and variance of the observed data, respectively. With a Gaussian prior distribution $p(s) = N(\eta, s_p) \equiv N_{pri}$, the posterior distribution is also of a Gaussian form.

$$p(s|o^n) = \frac{p(o^n|s)p(s)}{p(o^n)} = N(\eta_{post}, s_{post}) \equiv N_{post}, \tag{13}$$

where $\eta_{post} = \frac{n s_p \bar{o} + s_L \eta}{n s_p + s_L}$ and $\eta_{post} = \frac{n s_p \bar{o} + s_L \eta}{n s_p + s_L}$. The evidence $p(o)$ is a marginal likelihood:

$$p(o^n) = \int_{-\infty}^{\infty} p(o^n|s)p(s)ds = \sqrt{\frac{s_L}{n s_p + s_L}} \left(\frac{1}{\sqrt{2\pi s_L}}\right)^n \exp\left[-\frac{n}{2(s_p + s_L)}\delta^2 - \frac{n(s_p + s_L)}{2 s_L (n s_p + s_L)} V\right] \equiv e(\delta) \tag{14}$$

where $\delta = \eta - \bar{o}$ is a prediction error. The evidence is an inverse exponential function of the square of the prediction error. Hence, we describe the Gaussian evidence as $e(\delta)$. The evidence exponentially decreases as the prediction error increases. Surprise, the lower bound of free energy, is a negative log function of evidence, i.e., $-\log p(o^n) = -\log e(\delta)$. Thus, the surprise is a quadratic function of a prediction error.

The free energy of $n$ observations randomly obtained from a stimulus source following a Gaussian distribution of variance $s_L$ is formed as a quadratic function of the prediction error with coefficients of variance (for the derivation, see (Yanagisawa, 2021)):

$$F = A_F \delta^2 + B_F, \tag{15}$$

where the coefficients are functions of uncertainties, i.e., $A_F = \frac{1}{2} \frac{n}{n s_p + s_l}$ and $B_F = \frac{1}{2}\{\ln(n s_p + s_l) + (n-1)\ln s_l + n\ln 2\pi + nV/s_l\}$. To simplify further analysis, we consider the case of a single data observation, $n$=1. When $n$=1, the coefficients $A_F$ and $B_F$ are simplified as $A_F = \frac{1}{2}\frac{1}{s_p + s_l}$ and $B_F = \frac{1}{2}\{\ln(s_p + s_l) + \ln 2\pi\}$.

The gradient $A_F$ is the inverse of the sum of two variances. Thus, both uncertainties increase the sensitivity of the free energy to prediction error.

Using the same Gaussian model with a single data observation, we derive the information gains, KLD and BS, as quadratic functions of the prediction error with the coefficients of variance:

$$KLD = D_{KL}(p(s)||p(s|o)) = A_{KLD}\delta^2 + B_{KLD}, \tag{16}$$

where the coefficients are $A_{KLD} = \frac{s_p}{2 s_L (s_p + s_L)}$ and $B_{KLD} = -\ln\frac{s_p + s_l}{s_L} + \frac{s_p}{s_L}$ ; and

$$BS = D_{KL}(p(s|o)||p(s)) = A_{BS}\delta^2 + B_{BS}, \tag{17}$$

where the coefficients are $A_{BS} = \frac{s_p}{2(s_p + s_L)^2}$ and $B_{BS} = \ln\frac{s_p + s_l}{s_L} - \frac{s_p}{s_p + s_l}$.

The prediction error always increases both KLD and BS. We found that the observation variation $s_L$ always increases the gradient of both KLD and BS because the partial derivatives of the gradients are always negative, i.e., $\frac{\partial A_{KLD}}{\partial s_L} < 0$ and $\frac{\partial A_{BS}}{\partial s_L} < 0$. This signifies that the lower the observation uncertainty (i.e., the more precise the observation), the more susceptible the information gains (KLD and BS) are to prediction errors.

However, the effects of prediction uncertainty on the sensitivity of prediction errors are inversed between KLD and BS. We found that the prediction uncertainty increases sensitivity for KLD but decreases it for BS because the partial derivatives are $\frac{\partial A_{KLD}}{\partial s_p} > 0$ and $\frac{\partial A_{BS}}{\partial s_p} < 0$. Thus, the lower the prediction uncertainty, the more susceptible the information gains in recognition, i.e., KLD, and the more susceptible the information gains expected from Bayesian updating prior beliefs, i.e., BS.

To compare KLD and BS in the gradient of functions of prediction error, we derived the difference and found that KLD is always greater than BS because the coefficients are always positive.

$$KLD - BS = A_{KLD-BS}\delta^2 + B_{KLD-BS} > 0, \tag{18}$$

where $A_{KLD-BS} = \frac{s_p^2}{2s_L(s_p+s_L)} > 0$ and $B_{KLD-BS} = \frac{s_p^2}{s_L(s_p+s_L)} > 0$. Therefore, for any prediction error, the information gain in recognition is greater than that expected from updating the prior.

## 3.2 Convexity of information gain function by considering uniform noise

The Gaussian model suggests that prediction error always increases information gain. This is because the likelihood function is distributed over an infinite band and deviates from the prior distribution as the prediction error increases. However, the Laplace approximation is valid only around the mode, and there is no guarantee that the tail of the likelihood is infinitely distributed following a Gaussian distribution. The rate-coding hypothesis suggests that the likelihood function is coded by the distribution of the firing rates of neurons. A single stimulation fires specific neural populations but not all neurons. Neurons that do not receive external stimulation fire spontaneously (Raichle, 2006). The frequency of the spontaneous firing activity of neurons is lower than the millisecond-order frequency of stimuli-driven neural activity (Destexhe, Rudolph, & Paré, 2003). We infer that such spontaneous firing activity is independent of the neural activity evoked by sensory observations and provides no information about the cause of sensory stimuli (observation). To represent the influence of such independent and spontaneous neural activity, we added an independent uniformed likelihood with very small constant probability $\varepsilon$ to the observation-based likelihood (Jones, 2016).

$$p_\varepsilon(o|s) = p(o|s) + \varepsilon \tag{19}$$

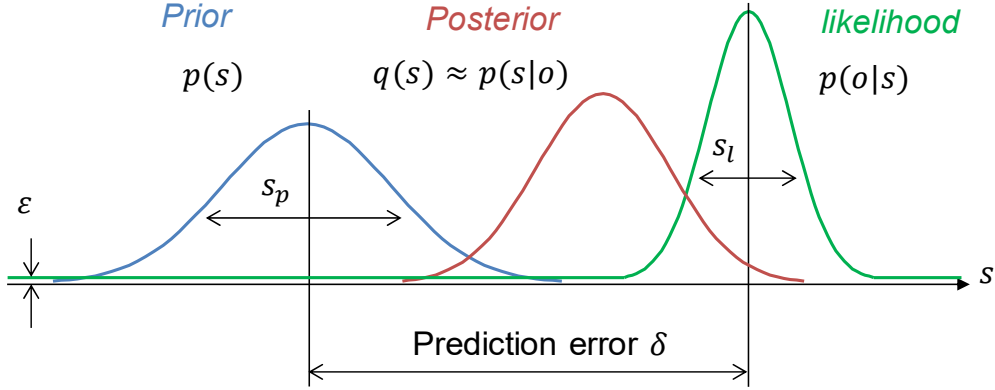

Fig. 3 Gaussian Bayesian model with uniform likelihood. $s_p$: prior variance, $s_l$: Gaussian likelihood variance, $\delta$: prediction error, and $\varepsilon$: probability of uniform likelihood.

This uniform likelihood addition flattens the tail of the Gaussian likelihood function, as shown in Fig. 3. The effect of the Gaussian tail becomes negligible as the prediction error increases. Therefore, we infer that adding a uniform likelihood is the simplest modeling method to represent the likelihood of spontaneous neural activity and to ignore the effect of the Gaussian likelihood tail.

The evidence with the likelihood function is the Gaussian evidence and the constant probability.

$$p_\varepsilon(o) = \int_{-\infty}^{\infty} p_\varepsilon(o|s)p(s)ds = e(\delta) + \varepsilon \tag{20}$$

Note that surprise increases monotonically with respect to the prediction error. We find that the posterior distributions with the likelihood function form a weighted linear model of the Gaussian posterior and prior.

$$p_\varepsilon(s|o) = \frac{p(s)p_\varepsilon(O|S)}{p_\varepsilon(o)} = \frac{e(\delta)N_{post} + \varepsilon N_{pri}}{e(\delta) + \varepsilon} = w_{post}N_{post} + w_{pri}N_{pri}, \tag{21}$$

where $w_{post} = \frac{e(\delta)}{e(\delta) + \varepsilon}$ and $w_{pri} = \frac{\varepsilon}{e(\delta) + \varepsilon}$ are the standardized linear weights. When the prediction error is small, the term $\varepsilon N_{pri}$ is negligible because $\varepsilon$ is very small compared to $e(\delta)$. In this case, the posterior is approximated to the Gaussian posterior, $p_\varepsilon(s|o) \approx N_{post}$. Thus, the prediction error increases both information gains, KLD and BS. By contrast, when the prediction error increases toward infinity, the evidence converges to zero, $\lim_{\delta \to \infty} e(\delta) = 0$, where $V = 0$, because the evidence is the inverse exponential function of the prediction error. In this case, the Gaussian posterior is negligible, and thus, the posterior is approximated to the Gaussian prior, $\lim_{\delta \to \infty} p_\varepsilon(s|o) = N_{pri}$. When the posterior is equal to the prior, both information gains, KLD and BS, are zero. Thus, in the case of a large prediction error, where $e(\delta)$ is very small compared to $\varepsilon$, and $\varepsilon N_{pri}$ is dominant in the posterior, the information gains decrease to zero as prediction error increases toward infinity. We use $\varepsilon = 10^{-3}$ for the following analysis.

The standardized linear weights $w_{post}$ and $w_{pri}$ represent the dominances of the Gaussian posterior and prior, respectively, in the mixed posterior distribution. Fig. 4 shows the dominances as functions of prediction error $\delta$. When the prediction error is zero or small, the Gaussian posterior is dominant. For a certain prediction error, the prior becomes dominant as the prediction error increases. Fig. 5 shows an example of posterior distributions switching over from Gaussian posterior dominance to prior dominance. In the switching over area of prediction errors, the Gaussian posterior and prior are mixed with certain weights, $w_{post}$ and $w_{pri}$.

Using the posterior function, we derived KLD and BS:

$$KLD = D_{KL}[p(s)||p(s|o)] = KLD_N + \ln\left(1 + \frac{\varepsilon}{e(\delta)}\right) - I, \tag{22}$$

where $KLD_N$ is a KLD using only the Gaussian likelihood, and $I$ is an improper integral:

$$I = \int_{-\infty}^{\infty} N_{pri} \ln\left(1 + \frac{\varepsilon N_{pri}}{e(\delta)N_{post}}\right) ds = \int_{-\infty}^{\infty} N_{Pri} \ln\left[1 + \varepsilon(2\pi s_l)^{\frac{n}{2}} \exp\left\{\frac{n(s-\delta)^2 + nV}{2s_l}\right\}\right] ds. \tag{23}$$

Using the KLD, we derived BS as

$$BS = D_{KL}[p(s|o)||p(s)] = \frac{1}{e(\delta)+\varepsilon}\left[e(\delta)\left\{BS_N - \ln\left(1 + \frac{\varepsilon}{e(\delta)}\right) + J\right\} - \varepsilon KLD\right], \tag{24}$$

where $BS_N$ is the $BS$ of using only the Gaussian likelihood, and $J$ is an improper integral:

$$J = \int_{-\infty}^{\infty} N_{post} \ln\left(1 + \frac{\varepsilon N_{pri}}{e(\delta)N_{post}}\right) ds = \int_{-\infty}^{\infty} N_{post} \ln\left[1 + \varepsilon(2\pi s_l)^{\frac{n}{2}} \exp\left\{\frac{n(s-\delta)^2 + nV}{2s_l}\right\}\right] ds. \tag{25}$$

Because improper integrals $I$ and $J$ could not be solved analytically, we used a computational approach for further analysis.

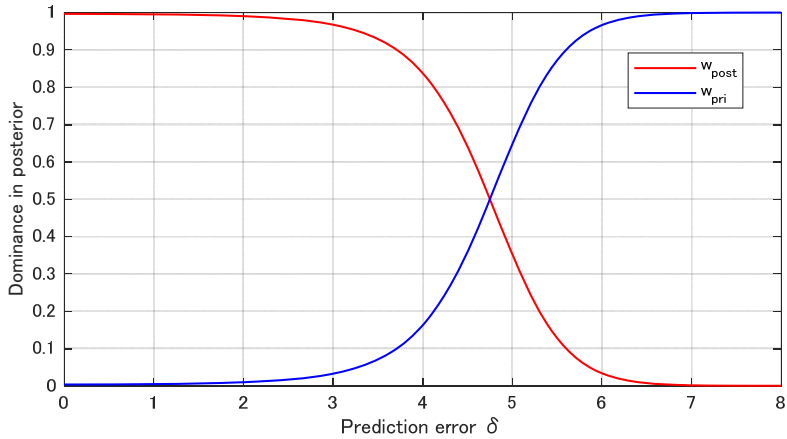

Fig. 4 Dominances of Gaussian posterior and prior in posterior distribution as functions of prediction error. The dominances swatch over at a certain prediction error level. (Variances: $s_p = 10.0$, $s_l$=1.0.)

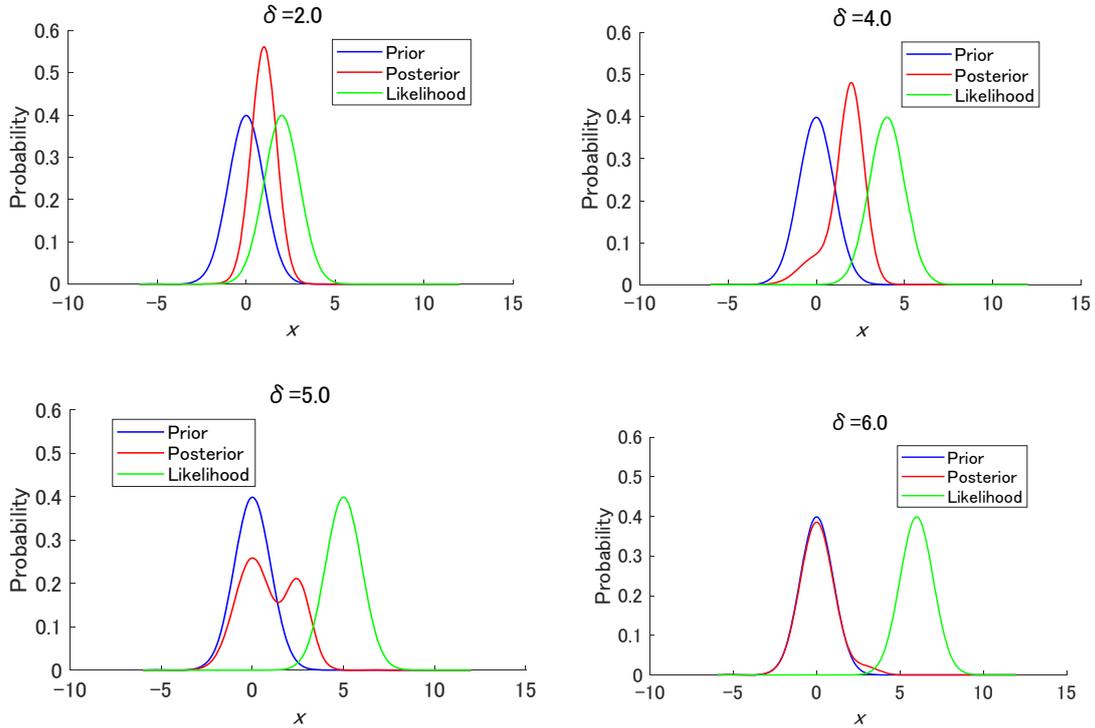

Fig. 5 Posterior distributions for different prediction errors ($\delta$=2.0, 4.0, 5.0, and 6.0). The dominances in the posterior distributions switch over from Gaussian posterior to prior. (Variances: $s_p = 10.0$, $s_l$=1.0.)

Fig. 6 (a) shows the information gains and their total value, $IG = KLD + BS$, as functions of the prediction errors. All information gains are upward-convex functions of the prediction errors. This convexity is general because when the prediction error is small, the Gaussian posterior is dominant in the posterior, and information gains increase as the prediction error increases; whereas when the prediction error is larger than a certain level, the prior becomes dominant, and the information gains decrease to zero as the prediction error increases.

Fig. 7 shows surprise as a function of the prediction error. Surprise increases monotonically with respect to the prediction error. Thus, the information gains are also upward-convex functions of surprise, and the total information gain $IG$ that induces positive emotions by reducing free energy is an upward-convex function of surprise (and prediction error). We infer that the upward-convex function of the total information gain represents the arousal potential function (i.e., the Wundt curve). Fig. 6 (b) shows an example of information gain as a function of surprise.

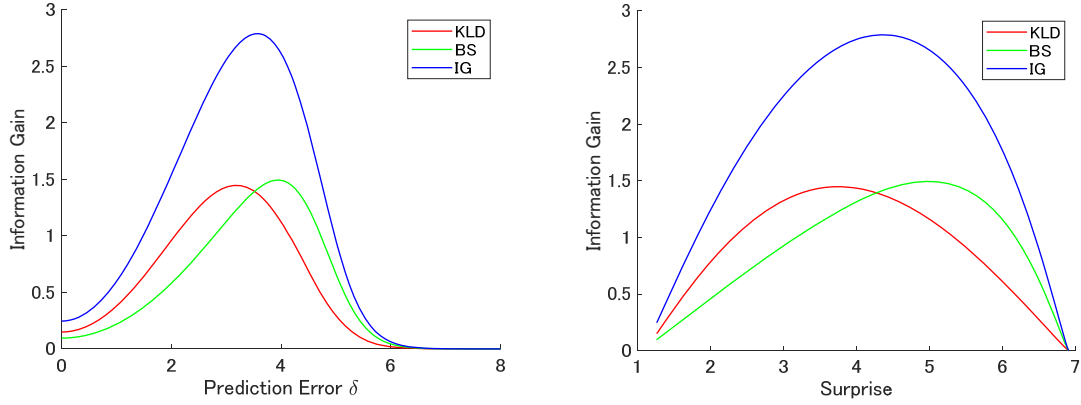

Fig. 6 Example of information gain functions of (a) prediction error and (b) surprise using Gaussian model with uniform noise. KLD and BS represent free energy reduction in recognition and prior updating (learning), respectively. Total information gain $IG$ is a summation of KLD and BS. (Uncertainties: $s_p = 10.0$, $s_l$=1.0.)

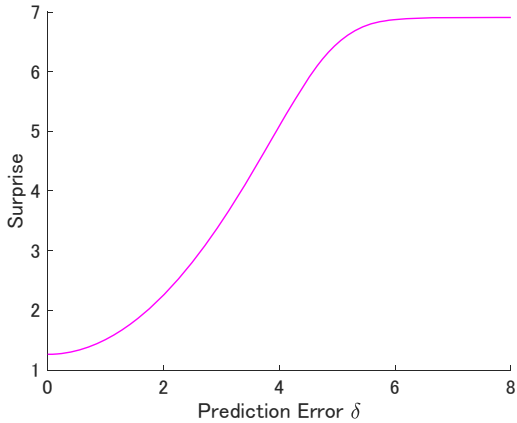

Fig. 7 Surprise as a function of prediction error. (variances: $s_p = 10.0$, $s_l$=1.0.)

Information gain functions are upward-convex and have a peak. We define the prediction errors that maximize information gains $KLD$, $BS$, and $IG$ as *optimal prediction errors* $\delta_{KLD}$, $\delta_{BS}$ and $\delta_{IG}$, respectively. Similarly, we define the surprises that maximize information gains $KLD$, $BS$, and $IG$ as *optimal surprises* $S_{KLD}$, $S_{BS}$, and $S_{IG}$, respectively. We use the term "optimal" because it represents the optimal arousal level that maximizes information gain (epistemic value) that evokes emotional valence. When the prediction errors are greater than $\delta_{KLD}$ and smaller than $\delta_{BS}$, $KLD$ and $BS$ have a negative relationship, where $KLD$ decreases as $BS$ increases, and vice versa. The prediction error that maximizes the total information gain $\delta_{IG}$ always falls into this area. Alternate maximizations of $KLD$ and $BS$ by decreasing and increasing the prediction error and surprise in this area iteratively reach the optimal surprise $S_{IG}$. This alternation generates fluctuations of surprise. The magnitude of fluctuation is determined by the difference between KLD and BS in the optimal prediction error $D_\delta = \delta_{BS} - \delta_{KLD}$ and surprise $D_S = S_{BS} - S_{KLD}$. In the next section, we analyze the effects of uncertainties on the optimal prediction errors and surprise, together with their differences.

### 3.3 Effects of uncertainties on information gains

The optimal prediction error and surprise change depending on uncertainties. We found the optimal prediction error and optimal surprise for all combinations of likelihood variances $s_l$ [1.0, 50] and prior variance $s_p$ [1.0, 50] in steps of 0.1 using the MATLAB fminbnd.m function, which is based on golden section search and parabolic interpolation.

Fig. 8 shows the maximum information gain as a function of the two uncertainties, $s_l$ and $s_p$. All maximum information gains decrease as $s_l$ increases, and increase as $s_p$ decreases. While $s_p$ approaches zero, the sensitivity of $s_l$ to the maximum information gains are low. The sensitivity of $s_l$ increases as $s_p$ increases. The peak of the maximum information gain is observed when $s_l$ is small, and $s_p$ is large. The maximum information gains of a large $s_l$ and large $s_p$ are greater than those of a small $s_l$ and small $s_p$. Fig. 9 shows examples of the maximum information gain as a function of $s_l$ and $s_p$. The maximum information gains increase exponentially as $s_l$ decreases. Thus, the sensitivity of $s_l$ to the maximum information gain increases as $s_l$ decreases. By contrast, the sensitivity of $s_p$ to information gain is significant when $s_p$ is small (e.g., from 1.0 to 10.0 in this example).

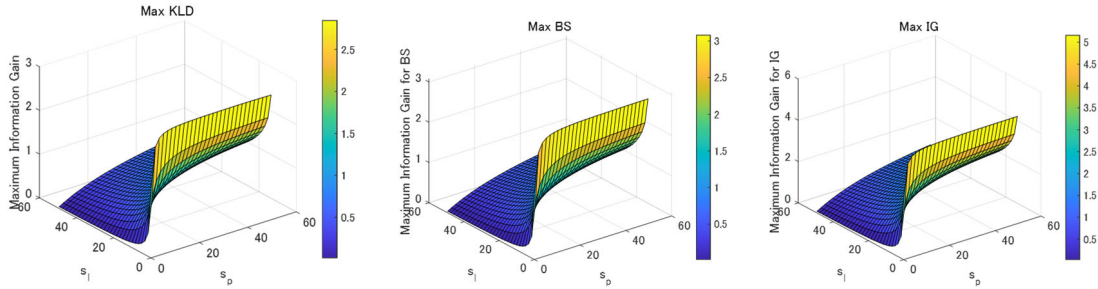

Fig. 8 Maximum information gains as function of uncertainties $s_l$ and $s_p$. (a) Max KLD, (b) Max BS, and (c) Max $IG$.

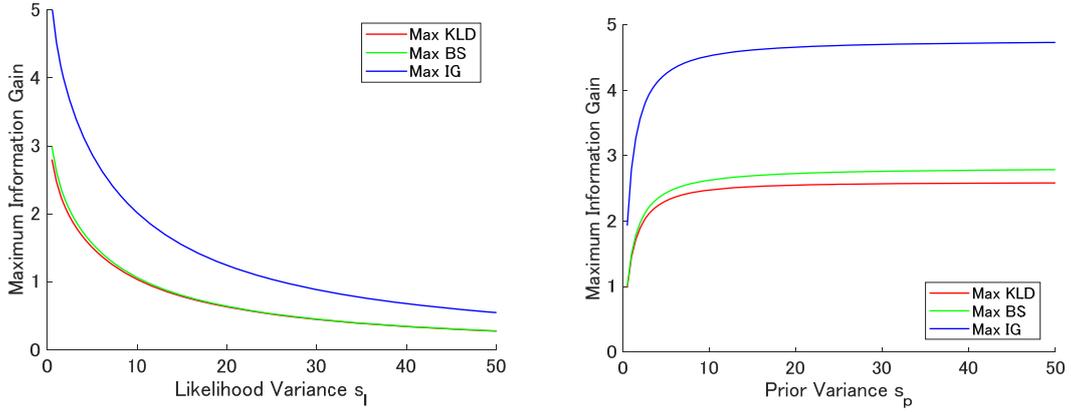

Fig. 9 Maximum information gains as functions of (a) likelihood variance when $s_p = 10$ and (b) prior variance when $s_l = 1.0$.

Fig. 10 shows the optimal prediction errors, $\delta_{KLD}$, $\delta_{BS}$ and $\delta_{IG}$, as functions of likelihood variance $s_l$ and prediction variance $s_p$. These two variances increase the optimal prediction errors. The sensitivity of $s_l$ is greater than that of $s_p$ in KLD. Fig. 11 shows an example of the optimal prediction error as a function of each uncertainty. All functions are monotonically increasing convex. $\delta_{KLD}$ is more sensitive to $s_l$ than $\delta_{BS}$. Thus, the difference $\delta_{KLD}$ and $\delta_{BS}$ decreases as $s_l$ increases. By contrast, $\delta_{KLD}$ is less sensitive to $s_p$ than $\delta_{BS}$. Thus, the difference increases as $s_p$ increases.

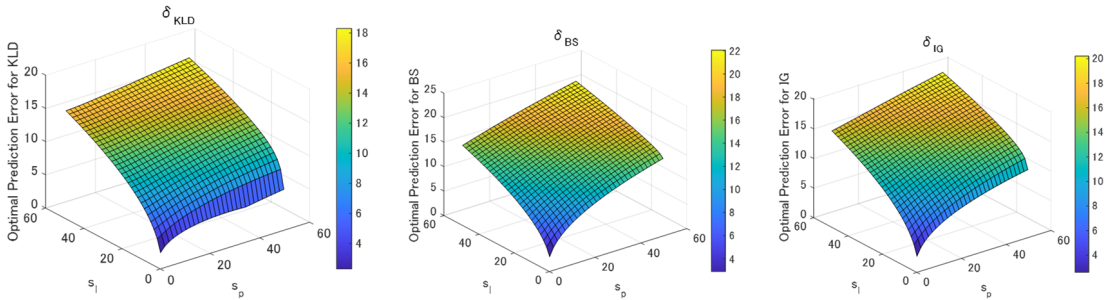

Fig. 10 Optimal prediction errors as functions of observation and prediction uncertainties for (a) $KLD$, (b) $BS$, and (c) $IG$.

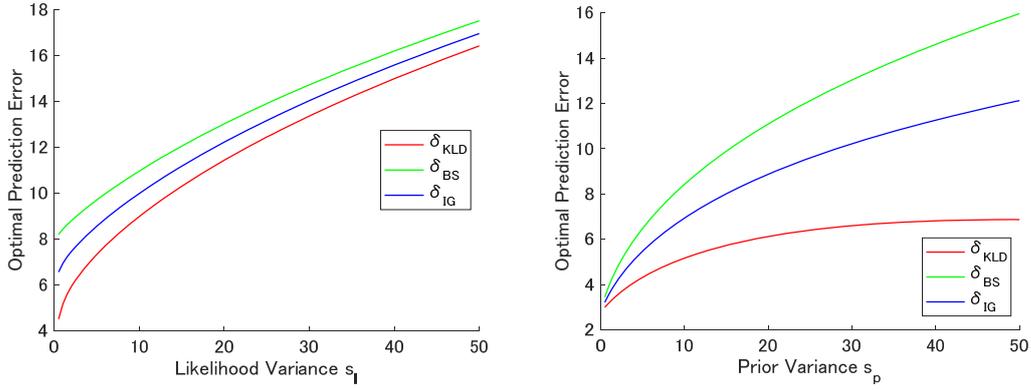

Fig. 11 Optimal prediction errors as functions of uncertainties, (a) likelihood variance when $s_p$=10.0 and (b) prediction variance when $s_l$=1.0.

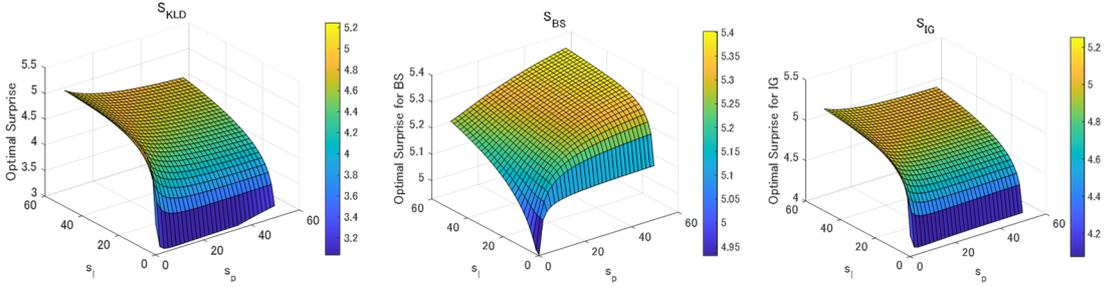

Fig. 12 Optimal surprises as functions of observation and prediction uncertainties for (a) KLD, (b) BS, and (c) $IG$.

Fig. 12 shows the optimal surprises $S_{KLD}$, $S_{BS}$, and $S_{IG}$ as functions of the two uncertainties. $s_l$ monotonically increases all optimal surprises. However, the effects of $s_p$ are different. $s_p$ decreases $S_{KLD}$ and increases $S_{BS}$. Fig. 13 shows examples of optimal surprises as functions of each uncertainty. $S_{KLD}$ is more sensitive to $s_l$ than $S_{BS}$, and thus, $S_{BS}$ approaches $S_{KLD}$ as $s_l$ increases. Consequently, the difference between $S_{KLD}$ and $S_{BS}$ decreases as $s_l$ increases. By contrast, $s_p$ decreases $S_{KLD}$ and increases $S_{BS}$. Thus, the difference between $S_{KLD}$ and $S_{BS}$ increases as $s_p$ increases.

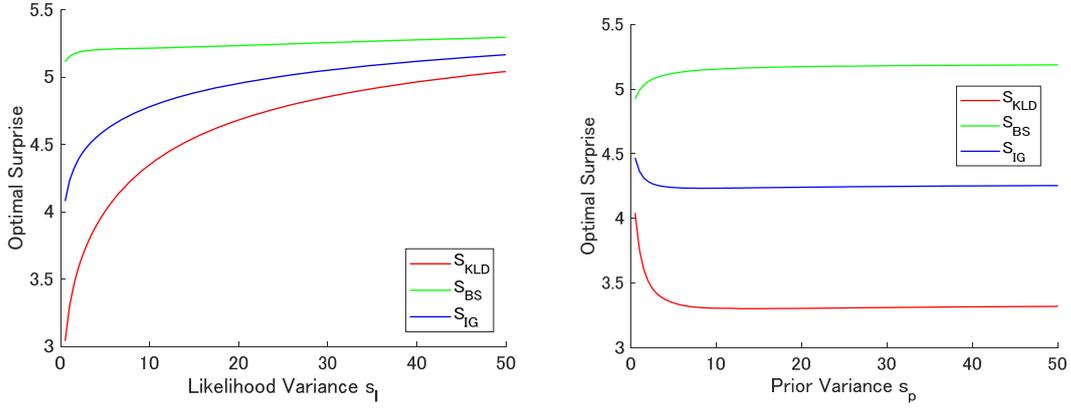

Fig. 13 Optimal surprises as functions of (a) likelihood variance when $s_p = 10.0$ and (b) prediction variance when $s_l = 1.0$.

Fig. 14 shows the differences in the optimal prediction error and optimal surprise. Both differences are always positive, and thus, $\delta_{BS} > \delta_{KLD}$ and $S_{BS} > S_{KLD}$. Both differences increase as $s_l$ decreases and $s_p$ increases. Thus, the larger the $s_p$, and the smaller the $s_l$, the larger the differences in both the optimal prediction errors and surprises. $s_l$ has the greatest sensitivity to increase the difference when $s_p$ is large.

For the optimal prediction errors, $s_p$ has the greatest sensitivity to increase the difference when $s_l$ is small. The difference in the optimal prediction error is larger when both $s_l$ and $s_p$ are large than when both $s_l$ and $s_p$ are small. By contrast, the difference in the optimal surprise is larger when both $s_l$ and $s_p$ are small than when both $s_l$ and $s_p$ are large.

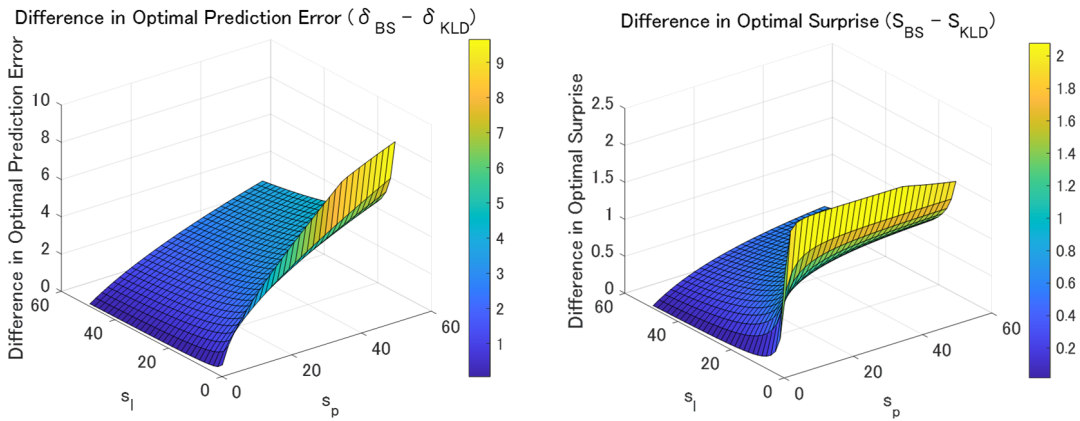

Fig. 14 Difference in optimal prediction error $D_\delta$ and difference in optimal surprise $D_S$.

## 4. Discussions

### 4.1 Arousal potential functions and curiosities

The results of the analysis using a Gaussian generative model with an additional uniform likelihood suggest that the two information gains, KLD and BS, form upward-convex functions of surprise and prediction errors (i.e., the distance between the prior mean and likelihood peak). The prediction error monotonically increases surprise, as shown in Fig. 7. Fig. 15 shows a schematic of the information gain functions that conceptualize the analytical results, as shown in Fig. 6 and the related emotions. Surprise, $-\ln p(o)$, corresponds to free energy minimized in recognition. A previous study argued that surprise represents arousal potential because minimized free energy consists of the summation of information content provided by novelty and perceived complexity, which are collative variables as dominant factors of arousal potential (Yanagisawa, 2021).

Berlyne suggested that an appropriate level of arousal potential induces a positive hedonic response, termed the optimal arousal level (Berlyne, 1960). Extreme arousal level caused by novel and complex stimuli may cause confusion. By contrast, a low arousal level with familiar and simple stimuli results in boredom. Thus, emotional valence shapes the upward-convex function of the arousal potential, termed the Wundt curve.

Berlyne also suggested that two epistemic curiosities, diversive and specific, drive the approach to the optimal arousal level (Berlyne, 1966). Diversive curiosity drives the pursuit of novelty, whereas specific curiosity drives the search for evidence of one's model predictions. Consequently, diversive curiosity increases the arousal potential to climb the Wundt curve on the left, from a low level of arousal (boredom). By contrast, specific curiosity motivates a decrease in the arousal potential to climb the Wundt curve on the right side from a high arousal level (confusion). The alternation between the two curiosity-driven activities approaches the optimal arousal level.

$KLD$ is a free energy reduction in recognition of a state $s$ given an observation $o$ that increases model evidence, $p(o) = \langle p(o|s) \rangle_{q(s)}$, where recognition $q(s)$ is updated from a prior $p(s)$ to true posterior $p(s|o)$. $BS$ is the expected information gain given by novel stimuli that corresponds to human surprise response to novelty (Itti & Baldi, 2009; Sekoguchi, Sakai, & Yanagisawa, 2019; Ueda et al., 2021; Yanagisawa et al., 2019). Therefore, we consider that specific curiosity drives an increase in KLD, whereas diversive curiosity drives an increase in BS.

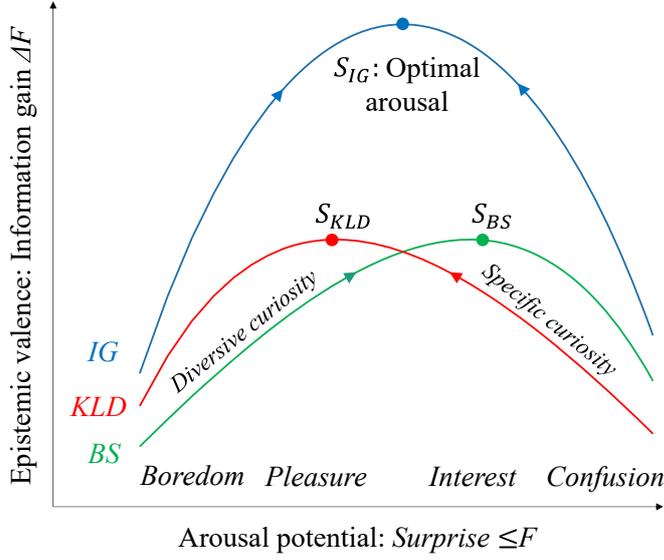

Fig. 15 Schematic of arousal potential functions and related emotions. Valence of epistemic emotions represented by information gains forms upward-function of arousal potential represented by free energy or surprise. Diversive and specific curiosity drive to maximize KLD and BS, respectively. These alternate maximizations achieve optimal arousal level with fluctuation of surprise. Emotions such as boredom, pleasure, interest, and confusion are induced by free energy and its fluctuations (see main text for detailed discussion)

## 4.2 Inquiry process and epistemic emotions

The analytical result shown in Fig. 14 demonstrate that the optimal surprise and optimal prediction error of BS is always greater than that of KLD, i.e., $S_{BS} > S_{KLD}$ and $\delta_{BS} > \delta_{KLD}$, respectively. This result suggests that maximizing information gain through novelty seeking (driven by diversive curiosity) requires a greater prediction error, causing greater surprise than that from maximizing information gain through evidence seeking (driven by specific curiosity).

When surprise is less than $S_{KLD}$, both KLD and BS monotonically increase as surprise increases. By contrast, when surprise is greater than $S_{BS}$, both KLD and BS monotonically decrease as surprise increases. Thus, the two curiosities increase and decrease prediction errors in the former and latter areas of surprise, respectively. However, when surprise is greater than $S_{KLD}$ and less than $S_{BS}$, KLD decreases, and BS increases as surprise increases. Thus, in this area of surprise, maximizing both the KLD and BS at same time is impossible. We infer that the two types of curiosity alternately maximize KLD and BS. This alternating maximization of information gains generates fluctuations of surprise. The optimal arousal level, as a maximum summation of KLD and BS, falls into this area. Therefore, the optimum arousal level, $S_{IG}$, involves fluctuations in surprise by alternately seeking novelty and evidence, driven by the two types of curiosity.

We consider that alternating the two kinds of curiosity by increasing and decreasing prediction errors represents an ideal inquiry process that achieves optimal arousal. This process provides continuous positive emotions through the continuous acquisition of maximum information gain (i.e., epistemic value). For example, "interest" is defined as disfluency reduction in fluency–disfluency theory (Graf & Landwehr, 2015). We previously formalized disfluency reduction as free energy reduction in recognition (i.e., KLD) from increased free energy (Yanagisawa et al., 2023). This corresponds to an increase in KLD from the high-surprise state shown in Fig. 15. Thus, "interest" is achieved by specific curiosity (i.e., climbing a hill of KLD from the right side in Fig. 15). By contrast, increasing KLD from the low-surprise state (i.e., climbing a hill of KLD from the left side in Fig. 15) may explain "pleasure" defined as increase in fluency (Graf & Landwehr, 2015). We have previously formalized fluency as KLD in recognition (Yanagisawa et al., 2023).

BS denotes the expected information gain, as discussed in the Methods section. Active inference suggests that an agent infers an optimal policy of action that minimizes expected free energy. The expected free energy includes the negative expected information gain as an epistemic value. This epistemic value drives curious behavior (Friston et al., 2017). Thus, diversive curiosity, formalized as maximizing the BS, corresponds to curiosity in active inference. We discuss the mathematical interpretations of KLD and BS in terms of the expected free energy in a later section.

## 4.3 Effect of uncertainties on optimal arousal level and epistemic values

We analyzed the effects of prediction and observation uncertainties, manipulated using prior and likelihood variances, on optimal information gains. Table 1 summarizes the effects of the two uncertainties in four quadrants for combinations of small and large uncertainties. A small prediction uncertainty $s_p$ indicates that the prior belief is certain because of, for example, prior experience and knowledge. However, prior beliefs are not always correct. The prediction error represents the error of prior belief from reality. Thus, a case with small $s_p$ and large prediction error indicates a preconceived notion. By contrast, a large $s_p$ denotes that the prior belief is uncertain, owing to, for example, a lack of prior knowledge and experience. Thus, observation uncertainty $s_l$ indicates precision of observations.

We evaluate the condition of uncertainties using four indices: maximum information gain ($\max IG$), optimal prediction errors ($\delta_{KLD}$, $\delta_{BS}$), optimal surprises ($S_{KLD}$, $S_{BS}$), difference in optimal prediction errors ($D_\delta$), and difference in optimal surprises ($D_s$). As shown in Fig. 8, the condition combining a small $s_l$ and large $s_p$ provides the largest $\max IG$ with the largest $D_\delta$ between small $\delta_{KLD}$ and moderate $\delta_{BS}$. A larger $D_\delta$ signifies a wider exploration range through alternations of diversive and specific curiosities. Smaller $S_{KLD}$ and $S_{BS}$ indicate less surprise as a cognitive load in the inquiry process. Therefore, the condition combining a small $s_l$ and large $s_p$ is the best solution to achieve the ideal inquiry process with the largest epistemic value (information gain; $\max IG$) and the largest range of exploration ($D_\delta$) under less cognitive load ($S_{KLD}$ and $S_{BS}$).

The condition combining a small $s_l$ and small $s_p$ is expected to yield the second largest epistemic value (information gain) under less cognitive load ($S_{KLD}$, $S_{BS}$); however, the range of exploration ($D_\delta$) is small. The condition combining a large $s_l$ and large $s_p$ is expected to result in a small information gain with a moderate range of exploration at the largest prediction error level. The condition combining a large $s_l$ and small $s_p$ is the worst case, corresponding to the smallest information gain and the smallest exploration range.

As overall trends, prediction uncertainty $s_p$ increases the range of exploration ($D_\delta$). This suggests that an extremely certain prior brief, such as a preconceived notion and strong assumption, suppresses the range of exploration, whereas an open mind involving a flat prior belief widens the range of exploration. The observation uncertainty $s_l$ decreases the expected maximum information gain (max $IG$). This suggests that precise observation increases expected information gains (epistemic value) with positive emotions. $s_l$ can be decreased in different ways; for example, by increasing the precision of stimuli, paying attention to stimuli, and improving the accuracy of the observation models.

Table 1 Summary of the effects of likelihood variance (observation uncertainty) $s_l$ and prior variance (prediction uncertainty) $s_p$ on maximum information gain, max $IG$, optimal prediction errors, $\delta_{KLD}$, $\delta_{BS}$, optimal surprises, $S_{KLD}$, $S_{BS}$, difference in optimal prediction errors, $D_\delta$, and difference in optimal surprises, $D_s$. $X: \Rightarrow Y$ signifies that $X$ dominantly affects $Y$. Solid and broken underlines denote positive and negative effects on epistemic emotions, respectively.

| $s_p: \Rightarrow D_\delta$ $s_l: \Rightarrow \max IG, D_s$ | Small $s_p$: small $D_\delta$ | Large $s_p$: |
|---|---|---|
| Small $s_l$: large $\max IG$ Small $\delta_{KLD}$ Small $S_{KLD}$, $S_{BS}$ large $D_s$. | Small $s_l$ and small $s_p$: Large $\max IG$, smallest $\delta_{KLD}$, $\delta_{BS}$, smallest $S_{KLD}$, $S_{BS}$, small $D_\delta$, large $D_s$. | Small $s_l$ and large $s_p$: Largest $\max IG$, small $\delta_{KLD}$, moderate $\delta_{BS}$, small $S_{KLD}$, $S_{BS}$, largest $D_\delta$, $D_s$. |
| Large $s_l$: small $\max IG$ large $S_{KLD}$ | Large $s_l$ and small $s_p$: smallest $\max IG$, moderate $\delta_{KLD}$, $\delta_{BS}$, largest $S_{KLD}$, moderate $S_{BS}$, smallest $D_\delta$, $D_s$. | Large $s_l$ and large $s_p$: small $\max IG$, largest $\delta_{KLD}$, $\delta_{BS}$, large $S_{KLD}$, largest $S_{BS}$, moderate $D_\delta$, $D_s$. |

## 4.4 Expected free energy and information gains

An active inference framework suggests that an agent's action policy is selected to minimize expected free energy (Friston et al., 2017; Parr et al., 2022; Smith, Friston, & Whyte, 2022). Here, we discuss the relationship between the expected free energy and the two types of information gains, KLD and BS, as drivers of specific and diversive curiosity, respectively.

Before giving observations by action, an agent calculates expected free energy under a policy $\pi$.

$$G_\pi = \langle \ln q(s|\pi) - p(s,o|\pi) \rangle_{q(S,O|\pi)}, \tag{26}$$

where $q(s,o|\pi) = q(s|\pi)p(o|s)$. This definition implies that the expected free energy is the free energy averaged by likelihood $p(o|s)$ of observations expected by future action under a policy $\pi$. The expected free energy forms the prior distribution of the policies. A policy is randomly selected based on the policy prior $p(\pi) = \sigma(-\gamma G_\pi)$ such that the expected free energy is minimized, where $\sigma(E)$ is a softmax function that transforms from energy $E$ to probability, and $\gamma$ is the precision of policy representing the *confidence* of policy selection.

The expected free energy is expanded in two terms using the decomposition of a generative model $p(s,o|\pi) = p(s|\pi)p(o|s)$.

$$\begin{aligned} G_\pi &= \langle \ln q(s|\pi) - p(s|\pi) \rangle_{q(S,O|\pi)} - \langle \ln p(o|s) \rangle_{q(S,O|\pi)} \\ &= D_{KL}[q(s|\pi)||p(s|\pi)] - \mathbb{E}_{q(O|\pi)}[\langle \ln p(o|s) \rangle_{q(S|O,\pi)}] \\ &= D_{KL}[q(s|\pi)||p(s|C)] + \mathbb{E}_{q(O|\pi)}[U_\pi] \end{aligned} \tag{27}$$

The first term is a KL divergence from a state prior to a recognition density under a policy, $q(s|\pi)$. We assume that the state prior is given from the agent's preference $C$, $p(s|\pi) = p(s|C)$. A preference refers to a desired state expected to be achieved through actions based on policy selection. The KL divergence, termed *risk in state*, refers to the difference between the preferred state and the state expected by acting with a policy. A lower KL divergence indicates that the desired state is more likely to be achieved. The second term is *expected uncertainty*, which represents uncertainty averaged over expected observations. This term is called *ambiguity* because it is equivalent to the entropy of likelihood,

$$-\langle \ln p(o|s) \rangle_{q(S,O|\pi)} = \langle -\ln p(o|s) \rangle_{p(O|S)q(S|\pi)} = \langle H(o|s) \rangle_{q(S|\pi)}.$$

The expected uncertainty is decomposed into two terms using a conditional probability definition:

$$p(o|s) = \frac{q(O|\pi)q(S|O,\pi)}{q(S|\pi)}.$$

$$\begin{aligned} \mathbb{E}_{q(O|\pi)}[U_\pi] &= -\mathbb{E}_{q(O|\pi)}\big[\langle \ln p(o|s) \rangle_{q(S|O,\pi)}\big] \\ &= -\mathbb{E}_{q(O|\pi)}\big[\langle \ln q(s|o,\pi) - \ln q(s|\pi) + \ln q(o|\pi) \rangle_{q(S|O,\pi)}\big] \\ &= -\mathbb{E}_{q(O|\pi)}\big[D_{KL}[q(s|o,\pi)||q(s|\pi)]\big] - \mathbb{E}_{q(O|\pi)}[\ln q(o|\pi)] \end{aligned} \tag{28}$$

The first term of the expected uncertainty is a negative KL divergence from approximate posterior to prior averaged by expected observations with a policy $\pi$. This KL divergence corresponds to the Bayesian surprise, $BS_\pi$. Thus, this term signifies the expected information gain by prior updating using predicted observations under policy $\pi$. Note that the observation is not yet given, and $BS_\pi$ is averaged based on the predicted distribution under a policy, $q(o|\pi)$.

The second term is entropy under a policy. By definition, surprise is the sum of the negative KLD and free energy.

$$-\ln q(o|\pi) = -D_{KL}[q(s|\pi)||q(s|o,\pi)] + \langle \ln q(s|\pi) - \ln q(s,o|\pi) \rangle_{q(S|\pi)} \tag{29}$$

Thus, the entropy is a summation of the negative predictive KLD and the predicted free energy.

$$-\mathbb{E}_{q(O|\pi)}[\ln q(o|\pi)] = -\mathbb{E}_{q(O|\pi)}[D_{KL}[q(s|\pi)||q(s|o,\pi)]] + \mathbb{E}_{q(O|\pi)}[F_\pi] \qquad (30)$$

In summary, the expected free energy under a policy is the sum of the risk, predicted free energy, and negative predicted information gains.

$$G_\pi = Risk + pF_\pi - (pKLD_\pi + pBS_\pi) \qquad (31)$$

where

risk in state: $Risk = D_{KL}[q(s|\pi)||p(s|C)],$ \qquad (32)

predicted free energy: $pF_\pi = \mathbb{E}_{q(O|\pi)}[\langle \ln q(s|\pi) - \ln q(s,o|\pi) \rangle_{q(S|\pi)}],$ \qquad (33)

predicted KLD: $pBS_\pi = \mathbb{E}_{q(O|\pi)}[D_{KL}[q(s|o,\pi)||q(s|\pi)]],$ \qquad (34)

and predicted Bayesian surprise: $pKLD_\pi = \mathbb{E}_{q(O|\pi)}[D_{KL}[q(s|\pi)||q(s|o,\pi)]].$ \qquad (35)

When preference is expected to be fully satisfied by a policy, the predicted state equals the state prior, $q(s|\pi) \approx p(s|C)$. In this case, the risk term becomes zero. However, the expected free energy still remains. The remaining expected free energy is the predicted free energy minus the two predicted information gains.

$$G_\pi \approx pF_\pi - (pKLD_\pi + pBS_\pi) \qquad (36)$$

The remaining expected free energy is minimized by maximizing the predicted information gains, $pKLD_\pi + pBS_\pi$, and minimizing the predicted free energy. Therefore, the two types of expected information gains, $pKLD_\pi, pBS_\pi$, drive the agent's action based on the active inference framework. This corresponds to the expected drives of the two types of curiosity.

## 4.5 Limitations and further discussions

The analytical results are based on a Gaussian generative model. A Gaussian model was used to independently manipulate the prediction errors and uncertainties and analyze their effects on information gains. Although Laplace approximation and the principle of maximum entropy reasonably support the Gaussian assumption, true distributions can be more complex than Gaussian distributions. For specific applications with complex distributions, further analysis based on the method proposed in this study paper for specific applications with complex distributions.

This study focusses on emotions induced by epistemic values (epistemic emotions) such as curiosity and interest. However, emotions are affected by individual preference and appraisal of the situation against objectives(Ellsworth & Scherer, 2003). We may expand the emotion model to include such preference-based emotions by introducing the pragmatic value formalized as risk term in expected free energy(Parr et al., 2022). The model does not consider individual capacity to process information. Surprise (free energy) exceeding the capacity may affect negative emotions.

This study was limited to analyzing two types of information gain linked to epistemic emotions as functions of surprise in a context-independent manner. Epistemic emotions based on epistemic values, such as curiosity, can be observed based on the agent's behavior. Active inference, where an action policy is inferred to minimize the expected free energy, can be used to simulate agent behavior based on epistemic emotions in a specific context (Friston et al., 2017). As discussed, the expected free energy includes two types of information gain. In future studies, it will be necessary to accumulate evidence of the model predictions based on correspondence between agent simulations and actual human behavior in a variety of specific contexts.

## 5. Conclusion

This study mathematically formulated arousal potential functions of epistemic emotions, such as curiosity and interest, that drive inquiry processes, based on information gains. Decrements in free energy in Bayesian recognition and prior belief update correspond to two types of information gain, i.e., KLD and BS, respectively. Free energy reduction induces positive emotions by reducing surprise caused by prediction errors and uncertainty, which provide information gains (i.e., epistemic value). We demonstrated that the two types of information gain form upward-convex curve functions of surprise using a Gaussian generative model with a uniform noise likelihood, and defined epistemic emotions as information gains (or decrements of free energy). An analysis using the model exhaustively revealed the effects of prediction and observation uncertainties on the peak of information gain functions as the optimal arousal level. Specifically, the analytical results suggest that the greater the prediction uncertainty and the lower the observation uncertainty, the greater the information gained through a larger exploration range.

These results provide general and fundamental knowledge to increase the valence of epistemic emotions that facilitate the inquiry process because the model is deduced from the synthesis of free energy minimization as the first principle of the brain and the well-established arousal potential theory. Therefore, this model framework is applicable to diverse areas that deal with epistemic emotions and motivations, such as education, creativity, aesthetics, affective computing, and related cognitive sciences. Further studies are needed to accumulate empirical evidence for the principle-based model and understand the relationship between the inquiry process and emotions in diverse complex situations.

## Acknowledgements

This research was supported by Japan Society for the Promotion of Science (KAKENHI Grant Number 21H03528, Mathematical model development of emotion dimensions based on variation of uncertainty and its application to inverse problems).